%% file: main.tex
\documentclass[letterpaper, 10 pt, journal, twoside]{IEEEtran}
\usepackage{amsmath,amsfonts}
\usepackage{array}
\usepackage[caption=false,font=normalsize,labelfont=sf,textfont=sf]{subfig}
\usepackage{textcomp}
\usepackage{stfloats}
\usepackage{url}
\usepackage{verbatim}
\usepackage{graphicx}
\usepackage{multirow}
\usepackage{amsthm}

\usepackage{bm}
\usepackage[table]{xcolor}
\usepackage{nomencl}
\usepackage[linesnumbered,ruled,vlined]{algorithm2e}
\usepackage[hidelinks]{hyperref}
\newcommand{\citeblue}[1]{{\textcolor{blue}{\cite{#1}}}}

\usepackage{amsthm}
\usepackage{amssymb}
\usepackage[noend]{algpseudocode}
\DeclareMathOperator*{\argmax}{arg\,max}
\DeclareMathOperator*{\argmin}{arg\,min}
\usepackage{amsmath}
\usepackage{float}
\usepackage{tabularx}
\usepackage{booktabs}
\usepackage{tikz}

\makenomenclature
\def\BibTeX{{\rm B\kern-.05em{\sc i\kern-.025em b}\kern-.08em
    T\kern-.1667em\lower.7ex\hbox{E}\kern-.125emX}}
\usepackage{balance}

\newcommand{\batchSize}{\mathcal{M}_\text{curr}}
\newcommand{\greedyGuild}{\mathcal{X}_{\textit{G$^2$}}}
\newcommand{\greedyGuildF}{\mathcal{X}_{\textit{G$^2$f}}}
\newcommand{\greedyGuildB}{\mathcal{X}_{\textit{G$^2$}b}}
\newcommand{\informedset}{\mathcal{X}_{\widehat{f}}}

\newcommand{\impcount}{N_{\text{imp}}}
\newcommand{\costImpRate}{\Upsilon_\text{CCI}}
\newcommand{\histImpRate}{\Upsilon_\text{HCI}}
\newcommand{\noImpThreshold}{\theta_\text{thold}}

\newcommand{\noimpcount}{\textit{l}_\text{level}}
\newcommand{\guildsamples}{\mathcal{M}_{\greedyGuild}}
\newcommand{\informedsamples}{\mathcal{M}_{\informedset}}

\newcommand{\GuildF}{\mathcal{X}_{\text{GuILD}_f}}
\newcommand{\GuildB}{\mathcal{X}_{\text{GuILD}_b}}
\newcommand{\neighborQ}{\mathcal{Q}_\text{comNeigh}}
\usepackage{setspace}
\begin{document}
\title{Tree-Based Grafting Approach for Bidirectional Motion Planning with Local Subsets Optimization}
\author{Liding Zhang$^{1}$, Yao Ling$^{1}$,  Zhenshan Bing$^{2,1}$, Fan Wu$^{1}$, Sami Haddadin$^{1}$, ~\IEEEmembership{Fellow,~IEEE}, \\Alois Knoll$^{1}$, ~\IEEEmembership{Fellow,~IEEE}
\thanks{Manuscript received: January, 2, 2025; Revised March, 10, 2025; Accepted April, 10, 2025. This paper was recommended for publication by Editor Júlia Borràs Sol upon evaluation of the Associate Editor and Reviewers’ comments.
\textit{(Corresponding author: Zhenshan Bing.)}
} 
\thanks{$^{1}$ Liding Zhang, Yao Ling, Zhanshan Bing, Fan Wu, Sami Haddain, Alois Knoll are with the School of Computation, Information and Technology (CIT), Technical University of Munich, 80333 Munich, Germany
        {\tt\footnotesize liding.zhang@tum.de}}%
\thanks{$^{2}$Zhenshan Bing is also with the State Key Laboratory for Novel Software Technology and the School of Science and Technology, Nanjing University (Suzhou Campus), China.}
\thanks{Digital Object Identifier (DOI): see top of this page.}
}
\markboth{IEEE Robotics and Automation Letters. Preprint Version. Accepted April, 2025}
{Zhang \MakeLowercase{\textit{et al.}}: Tree-Based Grafting Approach for Bidirectional Motion Planning with Local Subsets Optimization} 
\maketitle
\begin{abstract}
Bidirectional motion planning often reduces planning time compared to its unidirectional counterparts. 
It requires connecting the forward and reverse search trees to form a continuous path. However, this process could fail and restart the asymmetric bidirectional search due to the limitations of lazy-reverse search. To address this challenge, we propose Greedy GuILD Grafting Trees (G3T*), a novel path planner that grafts invalid edge connections at both ends to re-establish tree-based connectivity, enabling rapid path convergence. G3T* employs a greedy approach using the minimum Lebesgue measure of guided incremental local densification (GuILD) subsets to optimize paths efficiently. Furthermore, G3T* dynamically adjusts the sampling distribution between the informed set and GuILD subsets based on historical and current cost improvements, ensuring asymptotic optimality. 
These features enhance the forward search's growth towards the reverse tree, achieving faster convergence and lower solution costs. 
Benchmark experiments across dimensions from $\mathbb{R}^2$ to $\mathbb{R}^8$ and real-world robotic evaluations demonstrate G3T*'s superior performance compared to existing single-query sampling-based planners.
A video showcasing our experimental results is available at: \href{https://youtu.be/3mfCRL5SQIU}{\textcolor{blue}{https://youtu.be/3mfCRL5SQIU}}.
\end{abstract}

\begin{IEEEkeywords}
Bidirectional-tree grafting, greedy local subsets, optimal motion planning, sampling-based path planning
\end{IEEEkeywords}

\input{sections/sec1_intro}
\input{sections/sec3_prob}

\input{sections/sec4_algori}
\input{sections/sec5_analysis}
\input{sections/sec6_experiment}
\input{sections/sec7_conclusion}

\bibliographystyle{IEEEtran}
\bibliography{bibliography}

\end{document}

%% file: sections/sec1_intro.tex
\section{Introduction}
\IEEEPARstart{R}{obot} motion planning focuses on finding collision-free paths between start and goal configurations while avoiding obstacles~\citeblue{gammell2021asymptotically}. In high-dimensional configuration spaces (\textit{$\mathcal{C}$-spaces}), sampling-based algorithms are popular in robotics for their simplicity, fast exploration, and robustness against \textit{the curse of dimensionality}.
~\citeblue{zhang2024review, Orthey2023AnnualReview}.
In single-query motion planning, unidirectional planners grow a single tree from either the start or goal state, while bidirectional planners grow two trees—one forward from the start and one backward from the goal~\citeblue{Nayak2022}. Bidirectional planners are generally more efficient in finding solutions, their performance is often hindered by edge checks connecting the two trees, particularly in narrow passages and uneven sampling densities. 
We propose a grafting approach to regionally reconnect invalid edges from both trees, expediting the path search. In this paper, we aim to solve the motion planning problem as quickly as possible.
\begin{figure}[t!]
    \centering
    \begin{tikzpicture}
    
    \node[inner sep=0pt] (russell) at (0.0,0.0)
    {\includegraphics[width=0.49\textwidth]{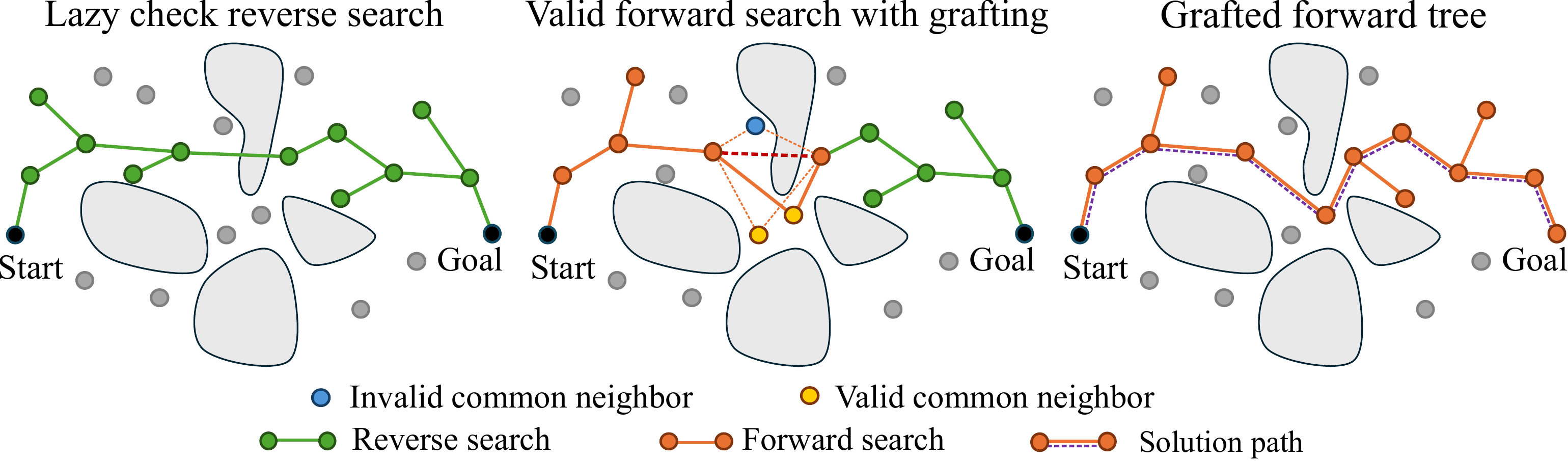}};
    \end{tikzpicture}
    \vspace{-1.5em} 
    \caption{Illustrates the grafting process in asymmetric bidirectional search. The forward search tree (orange) and the lazy-reverse search tree (green) grow independently from the start and goal configurations. Due to the limitations of lazy search, invalid edges can hinder the search process. Common neighbors are extracted and evaluated based on their heuristics to bypass obstacles (gray regions). Invalid common neighbors (blue) fail to bypass the obstacle, whereas valid common neighbors (yellow) successfully connect the two trees.}
    \label{fig: grafting}
    \vspace{-1.7em} 
\end{figure}
\subsection{Related Work}
Graph-based algorithms, such as Dijkstra's~\citeblue{dijkstra1959note}, explore all possible paths to compute the shortest route, while A*~\citeblue{hart1968formal} leverages heuristic functions to guide exploration. Sampling-based methods, including Rapidly-exploring Random Trees (RRT)~\citeblue{lavalle1998rapidly} and Probabilistic Roadmaps (PRM)~\citeblue{kavraki1996probabilistic} efficiently solve high-dimensional, multi-constrained path planning problems.
RRT grows a tree incrementally from the start toward the goal, where RRT-Connect~\citeblue{connect2000} enhances efficiency by growing trees from both ends. RRT*~\citeblue{karaman2011sampling} offering anytime solutions with guaranteed optimality by nearest neighbors~\citeblue{Zhang2024Elliptical} and rewiring. Informed RRT*~\citeblue{gammell2014informed} accelerates path optimization by limiting the search space via elliptical informed subsets~\citeblue{gammell2018informed}. 
Batch Informed Trees (BIT*)~\citeblue{gammell2015batch, gammell2020batch} integrates Informed RRT* and Fast Marching Trees (FMT*)~\citeblue{fmt2015}, advancing sampling techniques by grouping states (\textit{batch}) into compact implicit \textit{random geometric graphs} (RGGs)~\citeblue{penrose2003random}.
Adaptively Informed Trees (AIT*)~\citeblue{strub2022adaptively, strub2020adaptively} and Effort Informed Trees (EIT*)~\citeblue{strub2022adaptively} employ asymmetrical \textit{forward-reverse} search with sparse collision checks in the reverse search, enhancing exploration efficiency. Flexible Informed Trees (FIT*)~\citeblue{Zhang2024adaptive} utilize adaptive batch sizes for faster initial convergence. \textcolor{black}{Fully Connected Informed Trees (FCIT*)~\citeblue{Wilson2024fcit} utilize single instruction/multiple data parallelism to reduce the computational cost of edge checks.} However, those asymmetric bidirectional planners inefficiently restart the lazy-reverse tree when the forward tree encounters invalid edges. Our grafting method (Fig.~\ref{fig: grafting}) can utilize neighbor states of invalid edges to continue the forward search, thereby increasing the convergence rate.

The greedy informed set~\citeblue{Phone2022greedyinformed} improves convergence to the optimal by basing the informed set on the maximum admissible estimated cost of the path. Guided Incremental Local Densification (GuILD)~\citeblue{Scalise2023} further improves path optimization by refining the sampling process within local subsets of the problem domain. However, the informed set may converge slowly in complex environments with narrow passages. The greedy informed set and GuILD subsets are not asymptotic optimal, which can lead to suboptimal solutions. Furthermore, GuILD's dependence on carefully tuned beacon point selection increases the risk of getting trapped in local optima. To address these limitations, we propose a mathematically defined greedy beacon selector with a historically non-uniform distributed sampling strategy that maintains asymptotic optimality while incorporating greedy path optimization principles.
\subsection{Proposed Algorithm and Original Contribution}
This paper proposes Greedy GuILD Grafting Trees (G3T)*, a novel asymmetric bidirectional planner that accelerates motion planning in high-dimensional spaces. G3T* features a tree-based grafting method to reuse neighbor states for bridging invalid edges, avoiding costly tree restarts, and enhancing convergence in narrow passages. G3T* further employs beacon selection based on Lebesgue measures, efficiently targeting high-potential sampling regions to improve path quality and reduce computational overhead. This ensures robust performance in complex environments.
To maintain asymptotic optimality, G3T* uses a historical sampling strategy guided by the current and historical cost improvement metrics.
Experiments with state-of-the-art (SOTA) planners on cross-dimensional benchmarks and real-world robotic manipulation tasks showcase G3T*’s effectiveness in converging faster initial solutions (i.e., path cost) and enhanced path quality (i.e., path optimization).

The contributions of this work are summarized as follows:
\begin{enumerate}
    \item This paper presents a new tree-based grafting method to improve the convergence rate for bidirectional planners.
    \item The optimal selection of beacons and sampling regions via the Lebesgue measure for greedy GuILD subsets.
    \item Historical distributed sampling strategy based on cost improvements to ensure asymptotic optimality.
    \item Demonstrate the G3T* effectiveness across multiple domains with continuous goal regions in robotic systems.
\end{enumerate}

%% file: sections/sec3_prob.tex
\section{Problem Formulation and Preliminaries}
\subsection{Problem Definition}
\textit{The Optimal Path Planning Problem}~\citeblue{karaman2011sampling}: Given a planning problem with state space $\mathcal{X} \subseteq \mathbb{R}^n$, where $\mathcal{X}_{\text{obs}}$ denotes states in collision with obstacles, and $\mathcal{X}_{\text{free}} = cl(\mathcal{X} \setminus \mathcal{X}_{\text{obs}})$ represents permissible states, where $cl(\cdot)$ represents the \textit{closure} of a set. The initial state is $v_{\text{start}} \in \mathcal{X}_{\text{free}}$, and the desired final states are in $\mathcal{X}_{\text{goal}} \subset \mathcal{X}_{\text{free}}$. A continuous map $\xi: [0, 1] \mapsto \mathcal{X}$ represents a collision-free path, and $\Sigma$ is the set of all nontrivial paths.

The optimal solution, denoted as $\xi^*$, is the path minimizing a chosen cost function $c: \Sigma \mapsto \mathbb{R}{\geq 0}$. This path connects the initial state $v_{\text{start}}$ to any goal state $v_{\text{goal}} \in \mathcal{X}_{\text{goal}}$ through the free space:
\begin{equation}
\begin{split}
    \xi^* &= \argmin_{\xi \in \Sigma} \left\{ c(\xi) \middle| \xi(0) = v_{\text{start}}, \xi(1) \in v_{\text{goal}}, \right. \\
    &\qquad\qquad \left. \forall t \in [0, 1], \xi(t) \in \mathcal{X}_{\text{free}} \right\}.
\end{split}
\end{equation}
where $\mathbb{R}_{\geq 0}$ denotes non-negative real numbers. The cost of the optimal path is $c^*$.
Considering a discrete set of states, $\mathcal{X}_{\text{sample}} \subset \mathcal{X}$, represented as a graph with edges determined algorithmically by a transition function, we can characterize its properties using a probabilistic model, specifically implicit dense RGGs, i.e., $\mathcal{X}_{\text{sample}} = {v \sim \mathcal{U}(\mathcal{X})}$, as discussed in~\citeblue{penrose2003random}.

\begin{algorithm}[t]
\caption{Greedy GuILD Grafting Trees (G3T*)}
\SetKwInOut{Input}{Input}
\SetKwInOut{Output}{Output}
\SetKwFunction{sample}{sample}
\SetKwFunction{historicalSampling}{historicalSampling}
\SetKwFunction{grafting}{grafting}
\SetKwFunction{getRadius}{getRadius}
\SetKwFunction{lazyReverseSearch}{lazyReverseSearch}
\SetKwFunction{updateLazyReverseSearch}{updateLazyReverseSearch}
\SetKwFunction{forwardSearch}{forwardSearch}
\SetKwFunction{couldImproveForwardSearch}{couldImproveForwardSearch}
\SetKwFunction{isCollide}{isCollide}
\SetKwFunction{continueForwardSearch}{continueForwardSearch}
\SetKwFunction{terminateCondition}{terminateCondition}
\SetKwFunction{prune}{prune}

\DontPrintSemicolon
\small
\label{alg: g3t}
\Input{$\text{Start vertex}~v_{\text{start}}$, \text{goal region}~$\mathcal{X}_{\text{goal}}$}
\Output{$\text{Optimal feasible path}~\mathcal{T_F}$}

$\mathcal{X}_{\textit{sample}} \gets \{v_{\text{goal}}\}$, $E_\mathcal{F} \gets \emptyset$, $\mathcal{T_F} = (V_\mathcal{F}, E_\mathcal{F})$ \\

\While{\textbf{not} $\terminateCondition()$}{
    \textcolor{purple}{$\mathcal{X}_{\textit{sample}} \stackrel{+}{\leftarrow} \historicalSampling()$}\\
    $r \gets \getRadius()$\Comment{calculate RGG radius}\\
    $\mathcal{T_R}\gets\lazyReverseSearch()$\\
    \While{$\couldImproveForwardSearch(\mathcal{T_R})$}
    {   
        $E_\mathcal{F} \gets\forwardSearch(\mathcal{T_R})$\\

        \eIf{$\isCollide(E_\mathcal{F})$}{
        \textcolor{purple}{$\widehat{E}_\mathcal{P}\leftarrow\grafting(E_\mathcal{F}, r)$}\\
        \eIf{$\widehat{E}_\mathcal{P} \neq \emptyset$}{
            $\continueForwardSearch(\widehat{E}_\mathcal{P})$
        }{$\updateLazyReverseSearch()$}

    }{
    
    $\mathcal{T_F}\stackrel{+}{\leftarrow} E_\mathcal{F}$ \Comment{add the valid edges into the tree}\\
        
    }}
    $\prune(\mathcal{X}_{\textit{sample}})$
    }
\Return $\mathcal{T_F}$

\end{algorithm}

\subsection{Notation}~\label{subsec: notation}
The state space for the planning problem is denoted by $\mathcal{X} \subseteq \mathbb{R}^n$, where $n \in \mathbb{N}$. The start vertex is represented by $v_{\text{start}} \in \mathcal{X}$, and the goal states are denoted by $v_{\text{goal}} \in \mathcal{X}$. The set of sampled states is denoted by $\mathcal{X}_{\text{sample}}$. 
The forward and reverse search trees are denoted by $\mathcal{T}_\mathcal{F}$ and $\mathcal{T}_\mathcal{R}$, respectively. The vertices and edges in these trees, denoted by $V_\mathcal{F}$ and $V_\mathcal{R}$, correspond to valid states. The edges in the forward tree, $E_\mathcal{F} \subseteq V_\mathcal{F} \times V_\mathcal{F}$, represent valid connections between states. Each edge connects a source state, $v_s$, and a target state, $v_t$, denoted as $E(v_s, v_t)$. 
An admissible estimate of the cost of a path is denoted by $\hat{f}: \mathcal{X} \to [0, \infty)$.
For sets $A$, $B$, and $C$, where $B, C \subseteq A$, the notation $B \stackrel{+}{\leftarrow} C$ denotes the operation $B \leftarrow B \cup C$, and $B \stackrel{-}{\leftarrow} C$ denotes the operation $B \leftarrow B \setminus C$.

\textit{G3T*-specific Notation:}
The grafting edges set $E_\mathcal{P} \subset E_\mathcal{F}$, \textcolor{black}{which are defined by the common neighbor queue $\neighborQ$.} The optimal heuristic-sorted grafting edge is denoted as $\widehat{E}_\mathcal{P}$. The beacon vertex is represented as $v_\text{beacon} \in V_\mathcal{F}$. The front and back greedy GuILD subsets are denoted by $\greedyGuildF$ and $\greedyGuildB$, with the corresponding greedy costs denoted as $c_{\text{greedy}_f}$ and $c_{\text{greedy}_b}$, respectively. 
The greedy GuILD subsets are denoted as $\greedyGuild \subseteq \mathcal{X}$, where $\greedyGuild$ is a subset of the informed set $\informedset$.
The current and historical cost improvements are denoted by $\costImpRate$ and $\histImpRate$, respectively. The batch size of the current batch is denoted as $\batchSize$, with the distribution of $\guildsamples$ in the greedy GuILD and $\informedsamples$ in the informed set.

%% file: sections/sec4_algori.tex
\section{Greedy GuILD Grafting Trees (G3T*)}
\begin{algorithm}[t!]
\caption{{G3T*: Grafting}}
\label{alg:grafting}
\small
\DontPrintSemicolon
\SetKwIF{If}{ElseIf}{Else}{if}{}{else if}{else}{end if}%
\SetKwInOut{Input}{Input}
\SetKwInOut{Output}{Output}
\SetKwFunction{fullCheck}{fullCheck}
\SetKwFunction{foundValidEdgePair}{foundValidEdgePair}

\SetKwFunction{CalculateRatio}{calRatio}
\SetKwFunction{lazyCheck}{lazyCheck}
\SetKwFunction{nearest}{nearest}
\SetKwFunction{updateSparseCdRes}{updateSparseCdRes}

\SetKwFunction{expand}{expand}

\SetKwFunction{GetRadius}{getRNNRadius}
\SetKwFunction{bestEdgePair}{bestEdgePair}
\Input{($v_{\text{s}},v_{\text{t}}$) - The invalid edge, $r$ - RGG radius}
\Output{$\widehat{E}_\mathcal{P}$ - Valid edge-pair}

    $N_{\text{source}}\leftarrow\nearest(v_{\text{s}},r), N_{\text{target}}\leftarrow\nearest(v_{\text{t}},r)$\\
    $\neighborQ\leftarrow \emptyset$ \Comment{initialize common neighbor set}\\

    \ForEach{$v_{\text{i}} \in N_{\text{source}}$}  {
        \If{$v_{\text{i}} \in N_{\text{target}}$}
        {
        $\neighborQ\overset{+}{\operatorname*\leftarrow}v_{\text{i}}$\Comment{common neighbors of invalid-edge}
        }
        }
    \If{$\neighborQ\neq\emptyset$}
    {
    \ForEach{$v_{\text{i}} \in \neighborQ$}
        {
        \textcolor{purple}{$\widehat{E}_\mathcal{P} \leftarrow \bestEdgePair(\{(v_{\text{s}},v_{\text{i}}),(v_{\text{i}},v_{\text{t}})\})$}\Comment{Eq.~\ref{eq:HeuristicKey}}\\
        \If(\Comment{inexpensive edge-pair filter}){\textcolor{purple}{$\lazyCheck(\widehat{E}_\mathcal{P})$}}
            {   
                \If(\Comment{valid edge-pair}){$\fullCheck(\widehat{E}_\mathcal{P})$}{
                    $V_\mathcal{F}\leftarrow v_{\text{i}}$
                    \Comment{insert into forward tree}\\
                    $\mathcal{T_F}\overset{+}{\operatorname*{\leftarrow}}\expand(v_{\text{i}})$\Comment{add parent and children}\\
                    \Return $\widehat{E}_\mathcal{P}$
                }
            }
        }
    }

\Return $\textit{null}$
\end{algorithm}
In this section, we first introduce the concept of grafting. Next, we illustrate GuILD's greedy approach. Then, we propose the historical sampling. Finally, we analyze the probabilistic completeness and asymptotic optimality of G3T*.

%% file: sections/sec5_analysis.tex
\subsection{Bidirectional Tree-based Grafting Method}\label{subsec:elliptical knn}
The main idea behind our proposed grafting is to expand the source and target endpoints of invalid edges in the bidirectional tree search. First, \textcolor{black}{we obtain the common neighbor queue for both the source and target vertices} and then form two new edges as an edge-pair to continue the forward search.

Our algorithm is built on FIT*, which utilizes lazy checking in the reverse search to quickly explore the \textit{$\mathcal{C}$-spaces} (Alg.~\ref{alg: g3t}), while performing a full collision check in the forward search to ensure the validity of the path. We define the connection radius $r(m)$ during the construction of the RGGs. This defines the nearest neighbor samples and the density of the graph. \textcolor{black}{The radius decreases as the number of sampling vertices increases, leading to a denser RGG. Where $r(m)$ can be formulated as:
\begin{equation}
r(m):= \gamma\left(\frac{\log m}{m}\right)^{1/n},
\end{equation}
\begin{equation}
\gamma := 2\eta\left(1+\frac{1}{n}\right)^{\frac{1}{n}}\left(\frac{\min\{\lambda(\mathcal{X}_{\text{free}}),\lambda(\informedset)\}}{B_{1,n}}\right)^{1/n},
\end{equation}
here, 
$\gamma$ is a scaling factor tied to problem dimensions and the measure of the region, which defines the connectivity of RGGs~\citeblue{kiril2018con}. The term $\min\{\lambda(\mathcal{X}_{\text{free}}),\lambda(\informedset)\}$ ensures $\gamma$ is computable from $\mathcal{X}_{\text{free}}$ without an initial solution.} \(m\) represents the number of sampled states, \(\eta > 1\) is the RGG constant, \(n\) is the dimensionality of the problem domain, and \(\lambda(\cdot)\) denotes the Lebesgue measure (i.e., geometric volume). \(B_{1,n}\) represents the volume of the unit \(n\)-dimensional ball, $\informedset$ is the informed set.  
\textcolor{black}{In RGG, for any undirected graph $\mathcal{G}_n$, two vertices \(v\) and \(v'\) are said to be connected if there exists a path in $\mathcal{G}_n$ that links \(v\) to \(v'\). The connectivity of $\mathcal{G}_n$ is:
\begin{equation}
\lim_{m\to\infty}\mathbb{P}\left(\mathcal{G}_n~\text{is connected}\right):=\left\{\begin{array}{ll}0&\text{if }\gamma<\gamma^*\\1&\text{if }\gamma>\gamma^*\end{array}\right.,
\end{equation}
where  $\gamma^*=2(2n\cdot B_{1,n})^{-1/n}$. If every pair of vertices in the graph is connected, the graph is considered connected~\citeblue{kiril2018con}. In fully connected RGG, the nearest neighbor is the set of all vertices within a specified radius \(r(m)\) of a given vertex \(v\).} Mathematically, the neighborhood \(N(v)\) is expressed as:
\begin{equation}
N(v):=\{v'\in \mathcal{X}_{\textit{sample}}:\|v-v'\|_2\leq r(m)\},
\end{equation}
where \(\|v-v'\|_2\) denotes the distance between vertices \(v\) and \(v'\).
If a vertex \(v'\) is a neighbor of both \(v_s\) and \(v_t\), it is defined as a common neighbor of \(v_s\) and \(v_t\). The common neighbor queue \(\neighborQ\) of \(v_s\) and \(v_t\) is defined as: 
\begin{equation}
    \neighborQ(v_s,v_t)=\{v' \in \mathcal{X}_{\textit{sample}}\mid v'\in N(v_s)\cap N(v_t)\},
\end{equation}
%
%
%
the grafting method (Alg.~\ref{alg:grafting}) applies when the forward and reverse trees are connected in a bidirectional search. If the connecting edge $E(v_s, v_t)$ fails full collision detection (e.g., intersects an obstacle), the traditional asymmetric bidirectional search increases the lazy check resolution and restarts the reverse search. The new reverse tree is expanded from the goal and may have a different structure due to varying resolutions in lazy checking. It will continue attempting to connect to the source vertex of the invalid edge, $v_s$, making the search slower and more prolonged.
To address this limitation, the algorithm extracts the common neighbors of the search source vertex \(v_s\) and the target vertex \(v_t\) and sorts these common neighbors according to the search heuristics. Each common neighbor is evaluated to identify a valid edge-pair to bypass the obstacle. An edge-pair consists of two edges: \((v_s, v')\) and \((v', v_t)\). All edge-pairs form an edge queue \(E_\mathcal{P}\) described as:
\begin{equation}
 E_\mathcal{P} := \hspace{-0.4cm}\bigcup_{v' \in \neighborQ}\hspace{-0.4cm}\{(v_s, v'),(v', v_t)\},
\end{equation}
where \(v'\) represents a common neighbor of \(v_s\) and \(v_t\), which is lexicographically ordered based on the heuristic function value of its corresponding edge-pair using the heuristic key:
\begin{equation}
\text{key}_{\neighborQ}^{\text{G3T}^{*}}(v') :=
\begin{cases} 
\widehat{c}(v_s, v') + \widehat{c}(v', v_t), \\
\overline{e}(v_s, v') + \overline{e}(v', v_t),
\end{cases}
\end{equation}
here, $\widehat{c}$ represents the cost function used to evaluate the edge cost, and $\overline{e}$ is the heuristic effort required to validate the edge. By sorting the common neighbors $v'$ based on $\text{key}_{\neighborQ}^{\text{G3T}^{*}}(v')$, the algorithm efficiently identifies optimal edge-pairs that bypass obstacles and re-connect the forward and reverse trees.
\begin{equation}
\label{eq:HeuristicKey}
\begin{split}
\widehat{E}_\mathcal{P} &:=  \hspace{-0.6cm}\argmin_{\{(v_s, v'),(v', v_t)\} \in {E}_\mathcal{P}}\hspace{-0.6cm} \{  \text{key}_{\neighborQ}^{\text{G3T}^{*}}(v') | v'\in\neighborQ, \\
&\qquad\qquad\qquad\qquad\forall\neighborQ\in\mathcal{X}_\text{sample}\},
\end{split}
\end{equation}
where $\widehat{E}_\mathcal{P}$ represents the optimal edge-pairs. G3T* extracts the highest-priority common neighbor $v'$ from the common neighbor queue $\neighborQ$ to construct the edge-pairs ${E}_\mathcal{P}$. The validity of the edge-pair is first evaluated through a less computationally expensive lazy check, followed by the more costly full check (Alg.~\ref{alg:grafting}, lines 8-10). If the edge-pair passes the full check, it returns to continue the forward search, bypassing obstacles and connecting the forward and reverse trees. This approach accelerates pathfinding, and eliminates the need for restarting the reverse search.
\subsection{Greedy Approach for Beacon Selector in GuILD}\label{subsec:adaptive-batch}
\begin{algorithm}[t!]
\caption{{G3T*: Greedy GuILD Subsets}}
\label{alg:GuILD}
\small
\DontPrintSemicolon
\SetKwIF{If}{ElseIf}{Else}{if}{}{else if}{else}{end if}%
\SetKwInOut{Input}{Input}
\SetKwInOut{Output}{Output}
\SetKwFunction{IsValid}{isValid}
\SetKwFunction{UniformSample}{UniformSample}
\SetKwFunction{BeaconSelector}{BeaconSelector}
\SetKwFunction{DistributionSample}{DistributionSample}
\SetKwFunction{pathBetween}{pathBetween}
\SetKwFunction{hyperEllipsoid}{hyperEllipsoid}

\SetKwFunction{nearest}{nearest}
\SetKwFunction{updateSparseCdRes}{updateSparseCdRes}

\Input{$v_{\text{start}}$, $v_{\text{goal}}$, $\xi\text{ - Path of solution from start to goal}$}
\Output{Greedy measure subset $\greedyGuild$}


$\GuildF \leftarrow \mathcal{X}_{\text{free}},  \GuildB \leftarrow \mathcal{X}_{\text{free}}$\Comment{initialize GuILD sets}\\
\ForEach{$v_{\text{i}} \in \xi$}  
{   
    $\xi_{\text{$f_i$}} \leftarrow \pathBetween(v_{\text{start}}, v_{\text{i}})$\Comment{front-part path}\\
    $\xi_{\text{$b_i$}} \leftarrow \pathBetween(v_{\text{i}}, v_{\text{goal}})$\Comment{back-part path}\\
    
    $c_{\text{$f_i$}} \leftarrow c(\xi_{\text{$f_i$}}), \; c_{\text{$b_i$}} \leftarrow c(\xi_{\text{$b_i$}})$\Comment{calculate path costs}\\

    $\mathcal{X}_{\text{$f_i$}} \leftarrow \hyperEllipsoid(v_{\text{start}}, v_{\text{i}}, c_{\text{$f_i$}})$\\
    $\mathcal{X}_{\text{$b_i$}} \leftarrow \hyperEllipsoid(v_{\text{i}}, v_{\text{goal}}, c_{\text{$b_i$}})$\\
    $\lambda(\mathcal{X}_{v_{\text{i}}}) \leftarrow \lambda(\mathcal{X}_{\text{$f_i$}}) + \lambda(\mathcal{X}_{\text{$b_i$}})$\Comment{total Lebesgue measures}\\

    \If(\Comment{Eq.~\ref{eq:beaconSelect}}){\textcolor{purple}{$\lambda(\mathcal{X}_{v_{\text{i}}}) \leq \lambda(\GuildF)+\lambda(\GuildB)$}}
    {
        \textcolor{purple}{$v_{\text{beacon}} \leftarrow v_{\text{i}}$}\Comment{minimal measure beacon vertex}\\
        $\xi_{\text{f}} \leftarrow \xi_{\text{$f_i$}}, \xi_{\text{b}} \leftarrow \xi_{\text{$b_i$}}$\Comment{update minimal paths}\\
        $\GuildF \leftarrow \mathcal{X}_{\text{$f_i$}}, \GuildB \leftarrow \mathcal{X}_{\text{$b_i$}}$\Comment{update subsets}\\
    }
}
$c_{\text{greedy}_f} = 0$, $c_{\text{greedy}_b} = 0$\Comment{initialize greedy costs}\\
\ForEach(\Comment{get greedy cost for front-part path}){$v_{\text{j}} \in \xi_{\text{f}}$}  
{   
    $d(\text{$v_j$}) \leftarrow \|v_{\text{start}} - v_{\text{j}}\|_2 + \|v_{\text{j}} - v_{\text{beacon}}\|_2$\Comment{front-distance}\\

    \If(\Comment{Eq.~\ref{eq:greedyCostF}}){$d(\text{$v_j$}) \geq c_{\text{greedy}_f}$}
    {
        $v_{\text{f}} \leftarrow v_{\text{j}}$\Comment{set front-part greedy vertex}\\
        $c_{\text{greedy}_f} \leftarrow d(\text{$v_j$})$\\
        
    }
}
\ForEach(\Comment{get greedy cost for back-part path}){$v_{\text{k}} \in \xi_{\text{b}}$}
{ 
    $d(\text{$v_k$}) \leftarrow \|v_{\text{beacon}} - v_{\text{k}}\|_2 + \|v_{\text{k}} - v_{\text{goal}}\|_2$\Comment{back-distance}\\
    \If(\Comment{Eq.~\ref{eq:greedyCostB}}){$d(\text{$v_k$}) \geq c_{\text{greedy}_b}$}
    {
        $v_{\text{b}} \leftarrow v_{\text{k}}$\Comment{set back-part greedy vertex}\\
        $c_{\text{greedy}_b} \leftarrow d(\text{$v_k$})$\\
        
    }
}

$\greedyGuildF \leftarrow \hyperEllipsoid(v_{\text{start}}, v_{\text{beacon}}, c_{\text{greedy}_f})$\\
$\greedyGuildB \leftarrow \hyperEllipsoid(v_{\text{beacon}}, v_{\text{goal}}, c_{\text{greedy}_b})$\\
\textcolor{purple}{$\greedyGuild\leftarrow\mathcal{X}_{\text{free}} \cap\left(\greedyGuildF \cup \greedyGuildB\right)$}\Comment{update greedy GuILD sets}\\
\Return{$\greedyGuild$}

\end{algorithm}
In informed path planning, once the solution path $\xi$ is found, its cost $c(\xi)$ defines the informed set $\informedset$ for sampling and search. The informed set includes all paths with a cost less than $c(\xi)$, ensuring that any optimal path lies within it~\citeblue{gammell2018informed}. This reduces the search space and improves search efficiency by focusing on regions likely to contain optimal solutions:
\begin{equation}
\informedset =\left\{v\in\mathcal{X}_{\text{free}}\left|\left\|v-v_{\text{start}}\right\|_2+\left\|v_{\text{goal}}- v\right\|_2<c(\xi)\right\}\right.,
\end{equation}
where \(\|\cdot\|_2\) denotes the \(L_2\) norm (i.e., the Euclidean distance) between two vertices. The \(L_2\) informed set is defined as the intersection of the free space \(\mathcal{X}_{\text{free}}\) and an \(n\)-dimensional hyperellipsoid, which is symmetric along its transverse axis, \textcolor{black}{we define it as \textbf{S}ymmetric \textbf{H}yper\textbf{E}llipsoid (SHE):
\begin{equation}
\informedset := \mathcal{X}_{\text{free}} \cap\mathcal{X}_{\text{SHE}},
\end{equation}
\begin{equation}
\mathcal{X}_{\text{SHE}}:=\left\{v\in\mathbb{R}^n\left|\left\|v-v_{\text{start}}\right\|_2+\left\|v_{\text{goal}}- v\right\|_2<c(\xi)\right\}\right.,
\end{equation}
where the SHE of $n$-dimensional \textit{$\mathcal{C}$-space} ($\mathbb{R}^n$) is defined by the start vertex, the goal vertex, and the current path cost \(c(\xi)\).
The Lebesgue Measure (i.e., hypervolume) of the informed set in an \(n\)-dimensional Euclidean space \(\mathbb{R}^n\) is given by:
\begin{equation}
\lambda\left(\informedset\right)\leq\lambda\left(\mathcal{X}_{\text{SHE}}\right)=\frac{c(\xi)\left(c(\xi)^2-c_{\text{min}}^2\right)^{\frac{n-1}{2}}B_{1,n}}{2^n},
\end{equation}
where $B_{1,n}:={\pi^{\frac n2}}/{\Gamma\left(\frac n2+1\right)}$ is the Lebesgue measure of a $n$-dimensional unit ball, with the theoretical minimum cost $c_{\text{min}}:=\left\|v_{\text{goal}}- v_{\text{start}}\right\|_2$~\citeblue{gammell2018informed}.}
The informed set often provides limited improvements in sample efficiency, as its measure, $\lambda(\informedset)$, depends on the current solution cost. A high-cost solution may lead to a large informed set, potentially comparable to or larger than the full state space measure $\lambda(\mathcal{X})$, which slows convergence, increases computational costs, and reduces path optimization efficiency.
GuILD defines a vertex as a beacon vertex $v_{\text{beacon}}$, decomposing the planning problem into two smaller subproblems. This reduces the search domains and accelerates optimization convergence, partitioning the sampling space into two $n$-dimensional prolate hyperspheroids, $\GuildF$ and $\GuildB$.
\begin{figure}[t!]
    \centering
    \begin{tikzpicture}
    
    \node[inner sep=0pt] (russell) at (0.0,0.0)
    {\includegraphics[width=0.48\textwidth]{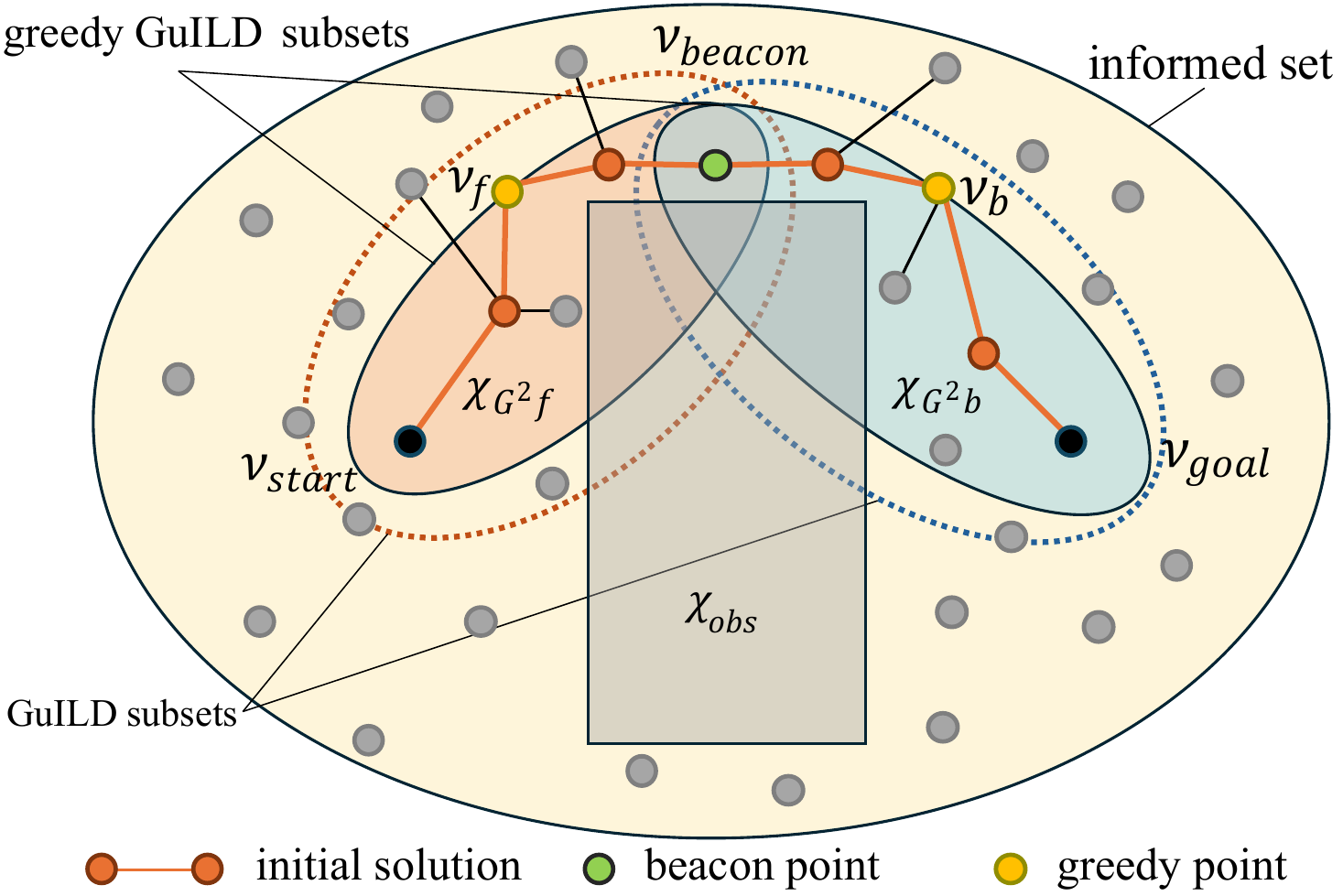}};
        
    \end{tikzpicture}
    \vspace{-1.em} 
    \caption{Illustrates the concept of GuILD and greedy GuILD. The problem domain is reduced to a smaller region by selecting a beacon point (green) along the solution path (orange). The first hyper-spheroid focuses on the start point and the beacon point, while the second hyper-spheroid focuses on the beacon point and the goal. Greedy vertices (yellow) are selected to evaluate cost values and enhance path optimization.}
    \label{fig: greedyguild}
    \vspace{-1.7em} 
\end{figure}
The greedy approach for beacon point selection $v_{\text{beacon}}$ in GuILD is defined via vertices along the solution path $\xi$ is selected as the beacon vertex that minimizes the sum of the Lebesgue measures of $\GuildF$ and $\GuildB$ (Alg.~\ref{alg:GuILD}):
\begin{equation}
\label{eq:beaconSelect}
v_{\text{beacon}} := \argmin_{v \in \xi} \{\lambda\left(\GuildF \right) + \lambda\left(\GuildB \right)\},
\end{equation}

$\GuildF$: Front GuILD is defined by the start vertex \(v_{\text{start}}\) and the beacon vertex \(v_{\text{beacon}}\), with the font-path cost \(c_f\) of the first hyperspheroid as its transverse diameter.

$\GuildB$: Back GuILD is defined by the beacon vertex \(v_{\text{beacon}}\) and the goal vertex \(v_{\text{goal}}\), with the back-path cost \(c_b\) of the second hyperspheroid as its transverse diameter.
\begin{equation}
\GuildF:= \{v\mid v\in\mathcal{X},\|v_{\text{start}}-v\|_{2}+\|v-v_{\text{beacon}}\|_{2}\leq c_f\},
\end{equation}
\begin{equation}
\GuildB:= \{v\mid v\in\mathcal{X},\|v_{\text{goal}}-v\|_{2}+\|v-v_{\text{beacon}}\|_{2}\leq c_b\},
\end{equation}
where the cost \(c_f\) is defined as the cost along the current solution path \(\xi\) from \(v_{\text{start}}\) to \(v_{\text{beacon}}\) and \(c_b\) is the cost from \(v_{\text{beacon}}\) to \(v_{\text{goal}}\), respectively. The combination is defined as the total path cost $c(\xi) := c_f + c_b$.
By employing the beacon vertex \(v_{\text{beacon}}\) and the path \(\xi\), the greedy GuILD subsets $\greedyGuild$ are defined as:
\begin{equation}
\greedyGuild :=  \mathcal{X}_{\text{free}} \cap\left(\greedyGuildF \cup \greedyGuildB\right),
\end{equation}

$\greedyGuildF$: Front greedy GuILD is defined by the start vertex \(v_{\text{start}}\) and the beacon vertex \(v_{\text{beacon}}\), with the greedy cost \(c_{\text{greedy}_f}\) (Alg.~\ref{alg:GuILD}, line 13-18) as its transverse diameter. The greedy cost \(c_{\text{greedy}_f}\) is determined by identifying the front greedy vertex \(v_f\) along the path \(\xi_f\), which connects \(v_{\text{start}}\) to \(v_{\text{beacon}}\). This vertex \(v_f\) maximizes the sum of the \(L_2\) heuristic values, comprising the cost-to-come from \(v_{\text{start}}\) and the cost-to-beacon to \(v_{\text{beacon}}\).

$\greedyGuildB$: Back greedy GuILD is defined by the beacon vertex \(v_{\text{beacon}}\) and the goal vertex \(v_{\text{goal}}\), with the greedy cost \(c_{\text{greedy}_b}\) as its transverse diameter. The \(c_{\text{greedy}_b}\) (Alg.~\ref{alg:GuILD}, line 19-23) is determined by the back greedy vertex \(v_b\) along the path \(\xi_b\), which connects \(v_{\text{beacon}}\) to \(v_{\text{goal}}\). This vertex \(v_b\) maximizes the sum of the \(L_2\) heuristic values, including the cost-to-come from \(v_{\text{beacon}}\) and the cost-to-goal to \(v_{\text{goal}}\) (Fig.~\ref{fig: greedyguild}), as follows:
\begin{figure}[t!]
    \centering
    \begin{tikzpicture}
    
    \node[inner sep=0pt] (russell) at (0.0,0.0)
    {\includegraphics[width=0.48\textwidth]{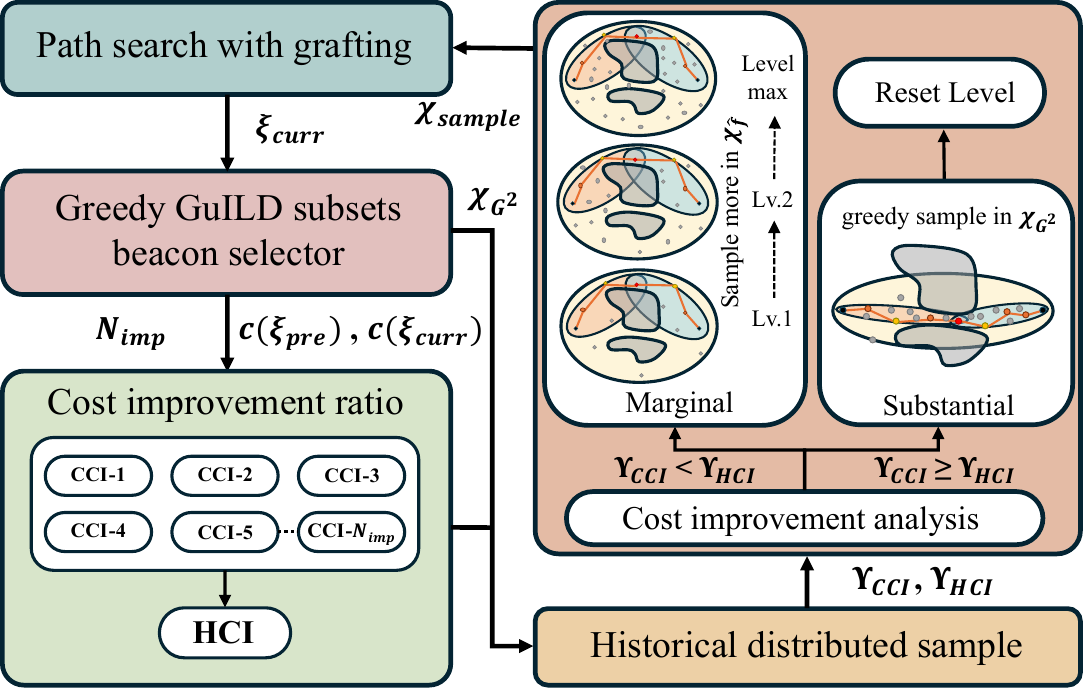}};
        
    \end{tikzpicture}
    \vspace{-0.7em} 
    \caption{Flowchart of G3T* integrating historical sampling into grafting and Greedy GuILD subsets. The current path \(\xi_\text{curr}\) is refined using samples \(\mathcal{X}_\text{sample}\) within adaptive subsets guided by a beacon selector. CCI is evaluated at each cost improvement, with HCI metrics guiding adaptive adjustments to sampling and level resets. The process dynamically updates the informed sampling regions and prioritizes substantial levels to accelerate optimization.}
    \label{fig: HistSample}
    \vspace{-1.7em} 
\end{figure}
\begin{equation}
\greedyGuildF:= \{v\in\mathbb{R}^n\mid \|v_{\text{start}}-v\|_{2}+\|v-v_{\text{beacon}}\|_{2}\leq c_{\text{greedy}_f}\},
\end{equation}
\begin{equation}
\greedyGuildB:= \{v\in\mathbb{R}^n\mid \|v_{\text{beacon}}-v\|_{2}+\|v-v_{\text{goal}}\|_{2}\leq c_{\text{greedy}_b}\},
\end{equation}
where greedy costs are defined as:
\begin{equation}
\label{eq:greedyCostF}
c_{\text{greedy}_f} := \argmax_{v \in \xi_f} \{ \|v_{\text{start}}-v\|_{2} + \|v-v_{\text{beacon}}\|_{2} \},
\end{equation}
\begin{equation}
\label{eq:greedyCostB}
c_{\text{greedy}_b} := \argmax_{v \in \xi_b} \{ \|v_{\text{beacon}}-v\|_{2} + \|v-v_{\text{goal}}\|_{2} \}.
\end{equation}

Compared to the traditional informed set, the greedy GuILD subsets $\greedyGuild$ reduce the search space, thereby improving convergence speed and sampling efficiency. However, this method alone prioritizes exploring local paths near the current beacon vertex while neglecting potentially better paths farther away. As a result, even when a globally shorter path exists, the algorithm may fail to discover it. This is its primary drawback.

To address this limitation, this paper proposes a cost-aware historical distributed sampling strategy. By adopting non-uniform sampling, the method ensures asymptotic optimality while maintaining the benefits of local search efficiency.


\subsection{Historical Distributed Sampling Strategy}\label{subsec: sample.}

\begin{algorithm}[t!]
\caption{{G3T*: Historical Distributed Sampling}}
\label{alg:sampling}
\small
\DontPrintSemicolon
\SetKwIF{If}{ElseIf}{Else}{if}{}{else if}{else}{end if}%
\SetKwInOut{Input}{Input}
\SetKwInOut{Output}{Output}
\SetKwFunction{uniformSample}{uniformSample}
\SetKwFunction{currCostImp}{currCostImp}
\SetKwFunction{histCostImp}{histCostImp}
\SetKwFunction{resetLevel}{resetLevel}

\Input{$c_{\text{prev}}$ - Cost of previous solution path, $c_{\text{prev}}$ - Cost of current solution path, $\batchSize$ - Batch size}
\Output{$\mathcal{X}_\text{sample}$ - Sampled states}

\If(\Comment{no initial path}){$c_{\text{prev}} = \infty$ \textbf{and} $ c_{\text{curr}} = \infty
$}
{    $\mathcal{X}_\text{sample}\xleftarrow{+}\uniformSample(\mathcal{X}_{\text{free}}, \batchSize)$
}
\ElseIf(\Comment{first initial path}){$c_{\text{prev}} = \infty$ \textbf{and} $ c_{\text{curr}} \neq \infty$}
{
    $\mathcal{X}_\text{sample}\xleftarrow{+}\uniformSample(\greedyGuild, \batchSize)$
}
\ElseIf(\Comment{improved path}){$c_{\text{prev}}\neq \infty$ \textbf{and} $ c_{\text{curr}}\neq\infty
$}
{
    $\impcount\leftarrow\ \impcount + 1 $\\
    
    $\costImpRate\leftarrow\currCostImp(c_{\text{prev}}, c_{\text{curr}})$\Comment{Eq.~\ref{eq:currentImp}}\\
    $\histImpRate\leftarrow\histCostImp(c_{\text{prev}}, c_{\text{curr}},\impcount)$\Comment{Eq.~\ref{eq:histImpRate}}\\

    \eIf(\Comment{marginal solution cost improvement}){$\costImpRate < \histImpRate$}
    {
        $\noimpcount\leftarrow\ \noimpcount + 1$\\
        \eIf(\Comment{$\noImpThreshold \in \mathbb{N^+}$ is level up threshold}){$\noimpcount\leq\noImpThreshold$}
        {   
             $\guildsamples\leftarrow\ \lfloor1/\noImpThreshold\rfloor \cdot\noimpcount\cdot \batchSize $\\


        }(\Comment{exceed threshold still can't improve})
        {
             $\guildsamples\leftarrow\ (\noImpThreshold\cdot\lfloor1/\noImpThreshold\rfloor) \cdot \batchSize $\\

        }
        $\mathcal{X}_\text{sample}\xleftarrow{+}\uniformSample(\greedyGuild, \guildsamples)$\\

    }(\Comment{substantial solution cost improvement})
    {
        $\mathcal{X}_\text{sample}\xleftarrow{+}\uniformSample(\greedyGuild, \batchSize)$\\

        $\noimpcount\leftarrow\resetLevel()$
    }
    $\informedsamples\leftarrow\ \batchSize - \guildsamples$\\
    $\mathcal{X}_\text{sample}\xleftarrow{+}\uniformSample(\informedset, \informedsamples)$\\
    
}

\Return $\mathcal{X}_\text{sample}$
\end{algorithm}
Sampling within the greedy GuILD subsets can accelerate convergence and optimization rates. However, this approach carries the risk of becoming trapped in a local optimum. In other words, vertices corresponding to potentially better solution paths might lie outside the greedy GuILD subsets \(\greedyGuild\). This paper proposes a cost-aware adjustment-based non-uniform distributed sampling strategy (Fig.~\ref{fig: HistSample}) to address this limitation. Adjust the batch size $\batchSize$ allocation between sampling within the \(\greedyGuild\) region and the area inside \(\informedset\) but outside \(\greedyGuild\), based on the \textit{historical cost improvement} (HCI) \(\histImpRate\) and the \textit{current cost improvement} (CCI) \(\costImpRate\). The batch size $\batchSize$ is defined as the number of valid vertices generated during each sampling process.
The concept of the current cost improvement ratio \(\costImpRate\) quantifies the extent to which the current solution improves relative to the previous one.
\begin{equation}
\label{eq:currentImp}
\costImpRate:=\frac{{c(\xi_{\text{prev}})-c(\xi_{\text{curr}})}}{c(\xi_{\text{prev}})},
\end{equation}

Moreover, the historical improvement rate \(\histImpRate\) is defined as the average value of all recorded cost improvement ratios:
\begin{equation}
\label{eq:histImpRate}
\histImpRate := \frac{1}\impcount\sum_{i=1}^{\impcount} \Upsilon_\text{CCI,i},
\end{equation}
where \(\impcount\) is the count of the number of cost improvements.
Uniform sampling is performed across the entire free space \(\mathcal{X}_{\text{free}}\)  when no initial path is found (Alg.~\ref{alg:sampling}, lines 1-2). Once an initial path is discovered, to accelerate the convergence of the current path, uniform sampling is restricted to the \(\greedyGuild\) region formed by the initial path (Alg.~\ref{alg:sampling}, lines 3-4). \textcolor{black}{After each subsequent update of the path, the sampler allocates sample vertices between the \(\greedyGuild\) 
region and the area outside \(\greedyGuild\) but within \(\informedset\), based on the values of HCI and CCI.}
\begin{equation} 
\limsup_{\impcount\to\infty}\mathbb{P}\left(\costImpRate > \Upsilon_{\text{HCI},{\impcount}}\right):=1,
\end{equation}
as the number of solution improvements \(\impcount\) approaches infinity, the historical improvement rate \(\histImpRate\) gradually approaches zero. This ensures that the current cost improvement ratio\(\costImpRate\) will always exceed the historical improvement rate \(\histImpRate\).
Then the marginal solution cost improvement level is reset (Alg.~\ref{alg:sampling}, line 18), ensuring that new sample vertices are generated within \(\informedset\). This mechanism guarantees asymptotic optimality.
The optimization speed and efficiency are improved while ensuring that the path-planning algorithm eventually converges to the optimal solution over time.

\subsection{Probabilistic Completeness and Asymptotic Optimality}\label{subsec: approx.}

Most informed sampling-based path-planning algorithms are probabilistically complete and asymptotically optimal, and G3T* upholds these properties. By leveraging uniform sampling, G3T* ensures that as the number of iterations $i$ approaches infinity, the entire \textit{$\mathcal{C}$-space} is  explored:
\begin{equation}
\lim_{i \to \infty} \mathbb{P} \left(\{V_\mathcal{F}\cup V_\mathcal{R}\} \cap \mathcal{X}_{\text{goal}} \neq \emptyset\right)= 1,
\end{equation}
this indicates that if a feasible path exists, G3T* will identify it, thereby guaranteeing probabilistic completeness.  

\textcolor{black}{By Lemmas 56, 71, and 72 in~\citeblue{karaman2011sampling}, the following holds in the limit of infinitely dense sampling:
\begin{equation}
\limsup_{q \to \infty} \mathbb{P} \left(\min_{\xi \in \Sigma_q} \left\{ c(\xi) \right\} = c^*\right) = 1,
\end{equation}
where $q$ is the number of samples, $\Sigma_q \subset \Sigma$ represents the set of valid paths identified by the planner, $c: \Sigma \rightarrow [0, \infty)$ is the cost function, and $c^*$ denotes the optimal cost. This indicates that G3T* guarantees asymptotic optimality, but only converges to the optimal path as the number of samples approaches infinity.}

%% file: sections/sec6_experiment.tex
\section{Experiments}\label{sec:Expri}
\begin{figure*}[t!]
    \centering
    \begin{tikzpicture}
    \node[inner sep=0pt] (russell) at (0,0)
    {\includegraphics[width=0.98\textwidth]{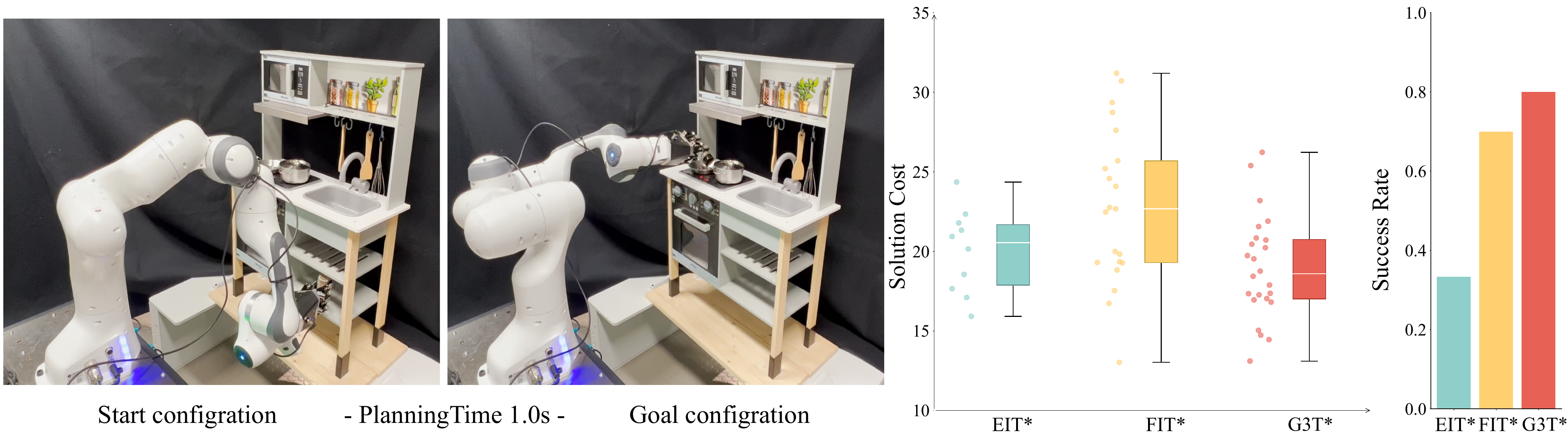}};

    \end{tikzpicture}
    \vspace{-0.7em} 
    \caption{Experimental results from Section~\ref{subsec:realExpri} are summarized above.  It illustrates the \textit{pepper} environment, where a manipulator grabs a pepper-box from the narrow bottom layer to the cooking section of the kitchen model. This scenario evaluates the planner's ability to address intrinsic challenges. The cost box plots represent solution costs for each planner, with white lines indicating mean cost progression (unsuccessful runs assigned infinite cost).}
    \label{fig: Realresult}
    \vspace{-1.7em}
\end{figure*}
\begin{figure}[t!]
    \centering
    \begin{tikzpicture}
    \node[inner sep=0pt] (russell) at (-4.0,0.0)
    {\includegraphics[width=0.24\textwidth]{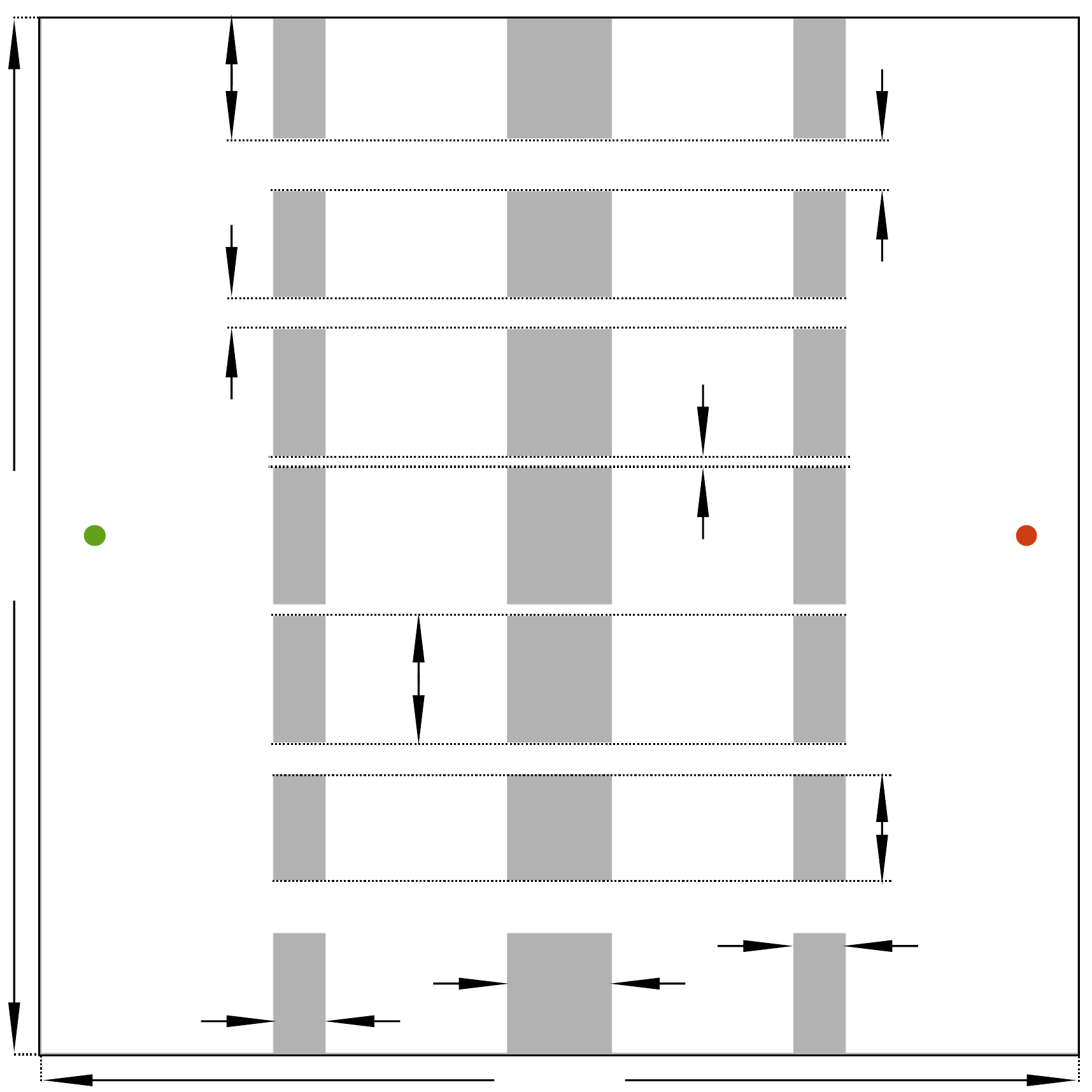}};
    \node[inner sep=0pt] (russell) at (0.25,0.0)
    {\includegraphics[width=0.24\textwidth]{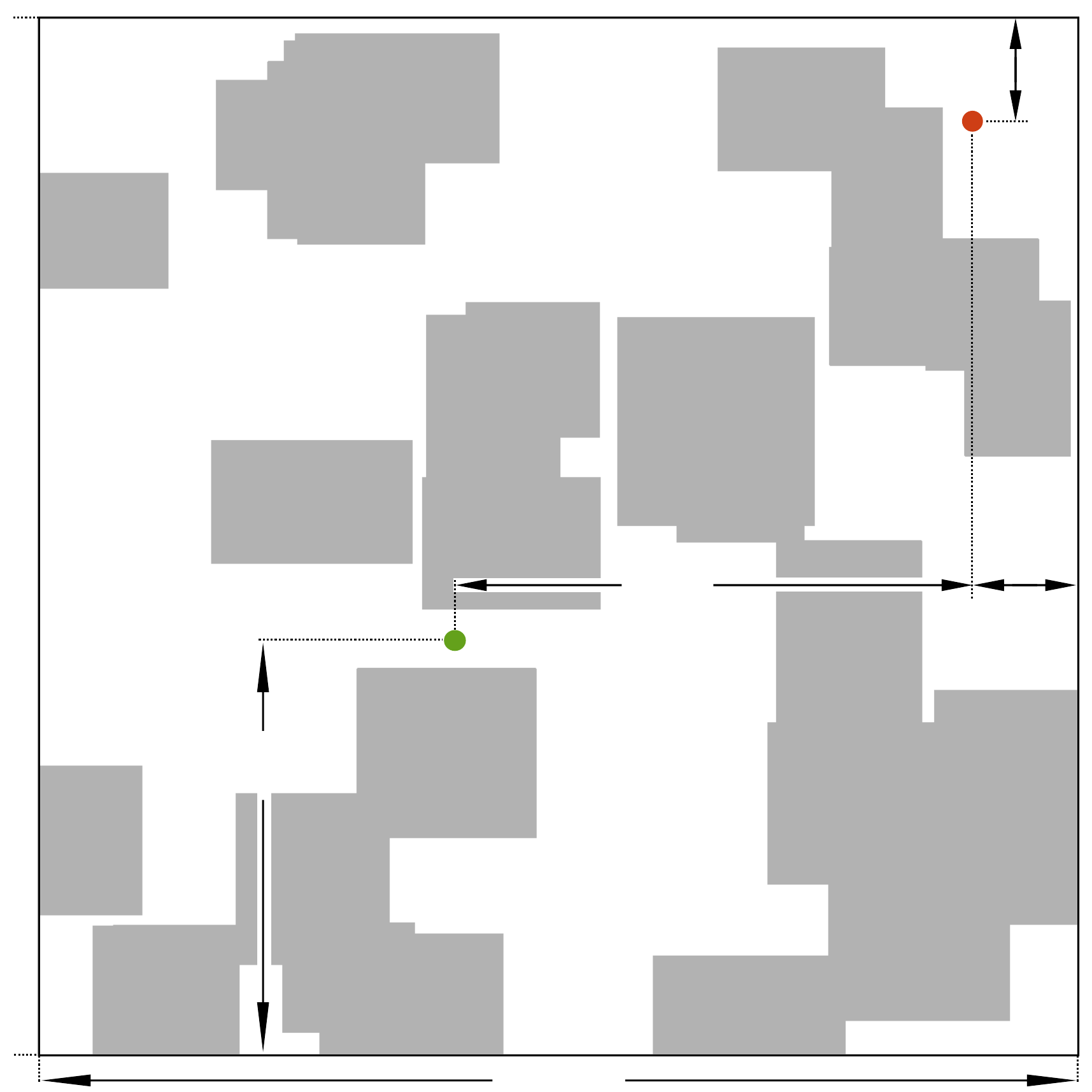}};
    \scriptsize
    
    \node [rotate=90] at (-5.4,0.93) {0.03};    
    \node [rotate=90] at (-4.68,-0.52) {0.125};    
    \node [rotate=90] at (-5.4,1.8) {0.12};
    \node [rotate=90] at (-2.53,-1.05) {0.1};
    \node [rotate=90] at (-3.55,0.6) {0.01};
    \node [rotate=90] at (-2.53,1.53) {0.05};
    \node at (-5.38,-1.72) {0.05};
    \node at (-4.38,-1.55) {0.1};
    \node at (-2.42,-1.72) {0.05};
    \node [rotate=90] at (-6.14,0.05) {1.0};
    \node at (-4.0,-2.15) {1.0};
    \node at (-5.53,-0.2) {(0.1,0.5)};
    \node at (-2.37,-0.2) {(0.9,0.5)};
    \node at (-5.6,0.3) {\color{teal} Start};
    \node at (-2.3,0.3) {\color{purple} Goal};
    \node at (0.74,-0.15) {0.5};
    \node at (0.25,-2.15) {1.0};
    \node [rotate=90] at (-0.89,-0.85) {0.4};
    \node [rotate=90] at (1.98,1.95) {0.1};
    \node  at (2.2,-0.3) {0.1};

    \node at (0.24,-0.36) {\color{teal} Start};
    \node at (2.1,1.5) {\color{purple} Goal};

    \node at (-4.0,-2.51) {\small \textcolor{black}{(a) Dividing Wall-gaps (DW)}};
    \node at (0.25,-2.51) {\small (b) Random Rectangles (RR)};
    \end{tikzpicture}
    \vspace{-1em} 
    \caption{Simulated planning problems were illustrated using a 2D representation, with state space \( \mathcal{X} \subset \mathbb{R}^n \) confined within a unit hypercube. Both the dividing wall-gaps and random rectangles benchmarks were evaluated across ten variations, and the corresponding results are shown in Fig.~\ref{fig: result}.
}
    \label{fig: testEnv}
    \vspace{-0.7em} 
\end{figure}
\begin{figure*}[t!]
    \centering
    \begin{tikzpicture}
    \node[inner sep=0pt] (russell) at (4.1,7.98)
    {\includegraphics[width=0.493\textwidth]{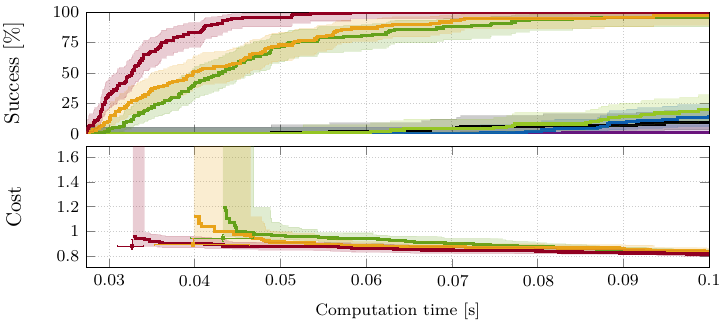}};
    \node[inner sep=0pt] (russell) at (4.1,3.54)
    {\includegraphics[width=0.493\textwidth]{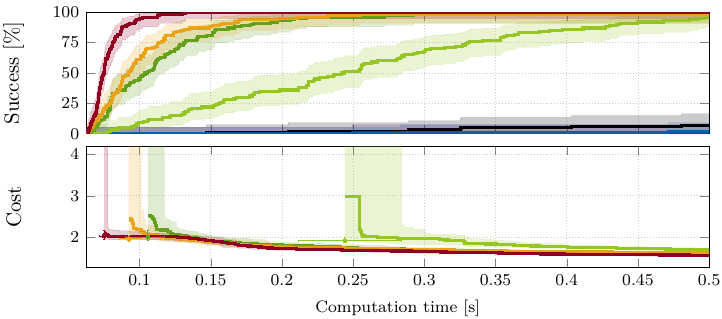}};
    \node[inner sep=0pt] (russell) at (4.05,-0.98)
    {\includegraphics[width=0.488\textwidth]{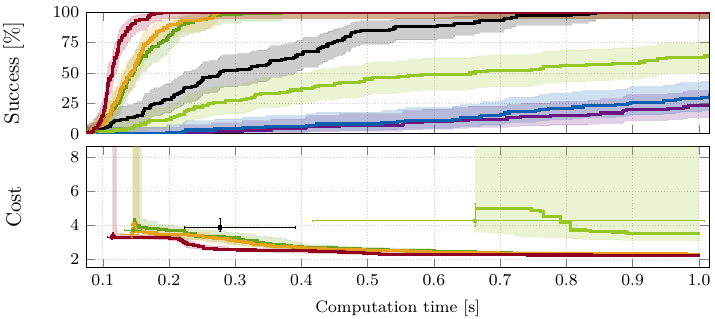}};

    \node[inner sep=0pt] (russell) at (-4.9,8)
    {\includegraphics[width=0.49\textwidth]{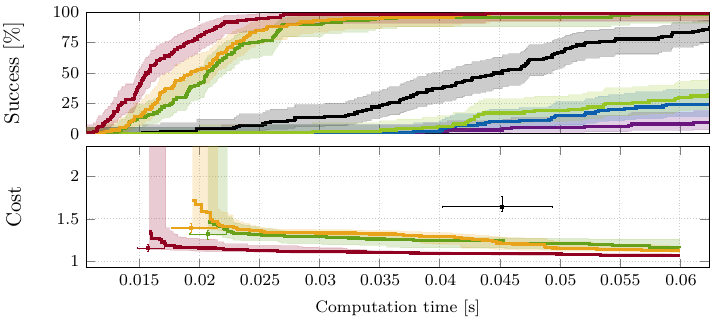}};  
    \node[inner sep=0pt] (russell) at (-4.9,3.5)
    {\includegraphics[width=0.49\textwidth]{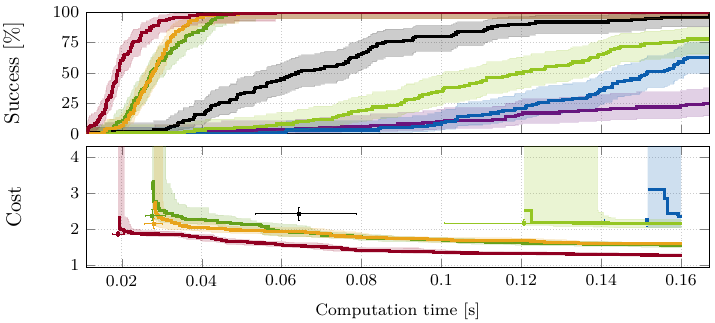}};
    \node[inner sep=0pt] (russell) at (-4.9,-1){\includegraphics[width=0.49\textwidth]{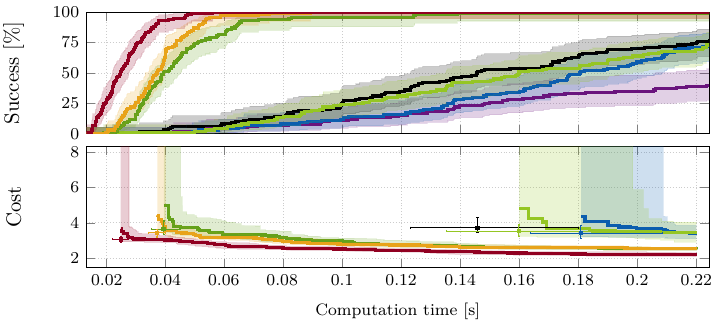}};

    \node[inner sep=0pt] (russell) at (0.0,-3.9){\includegraphics[width=0.75\textwidth]{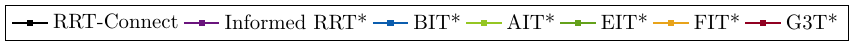}};

    \node at (-4.5,5.8) {\footnotesize \textcolor{black}{(a) Dividing Wall-gaps (DW) in $\mathbb{R}^2$ - MaxTime: 0.06s}};
    \node at (-4.5,1.3) {\footnotesize \textcolor{black}{(c) Dividing Wall-gaps (DW) in $\mathbb{R}^4$ - MaxTime: 0.16s}};
    \node at (-4.5,-3.2) {\footnotesize \textcolor{black}{(e) Dividing Wall-gaps (DW) in $\mathbb{R}^{8}$ - MaxTime: 0.22s}};

    \node at (4.5,5.8) {\footnotesize (b) Random Rectangles (RR) in $\mathbb{R}^2$ - MaxTime: 0.10s};
    \node at (4.5,1.3) {\footnotesize(d) Random Rectangles (RR) in $\mathbb{R}^4$ - MaxTime: 0.50s};
    \node at (4.5,-3.2) {\footnotesize(f) Random Rectangles (RR) in $\mathbb{R}^{8}$ - MaxTime: 1.00s};

    \end{tikzpicture}
    \vspace{-1.0em} 
    \caption{Detailed experimental results from Section~\ref{subsec:experi} are presented above. MaxTime is the planner's maximum allotted planning time. Fig. (a), (c), and (e) depict test benchmark dividing wall-gaps (DW) outcomes in $\mathbb{R}^2$ to $\mathbb{R}^{8}$, respectively. Panel (b) showcases random rectangle (RR) experiments in $\mathbb{R}^2$, while panels (d) and (f) demonstrate in $\mathbb{R}^4$ and $\mathbb{R}^{8}$. In the cost plots, boxes represent solution cost and time, with lines showing cost progression for optimal planners (unsuccessful runs have infinite cost). Error bars provide nonparametric 99\% confidence intervals for solution cost and time.}
    \label{fig: result}
    \vspace{-1.8em}
\end{figure*}
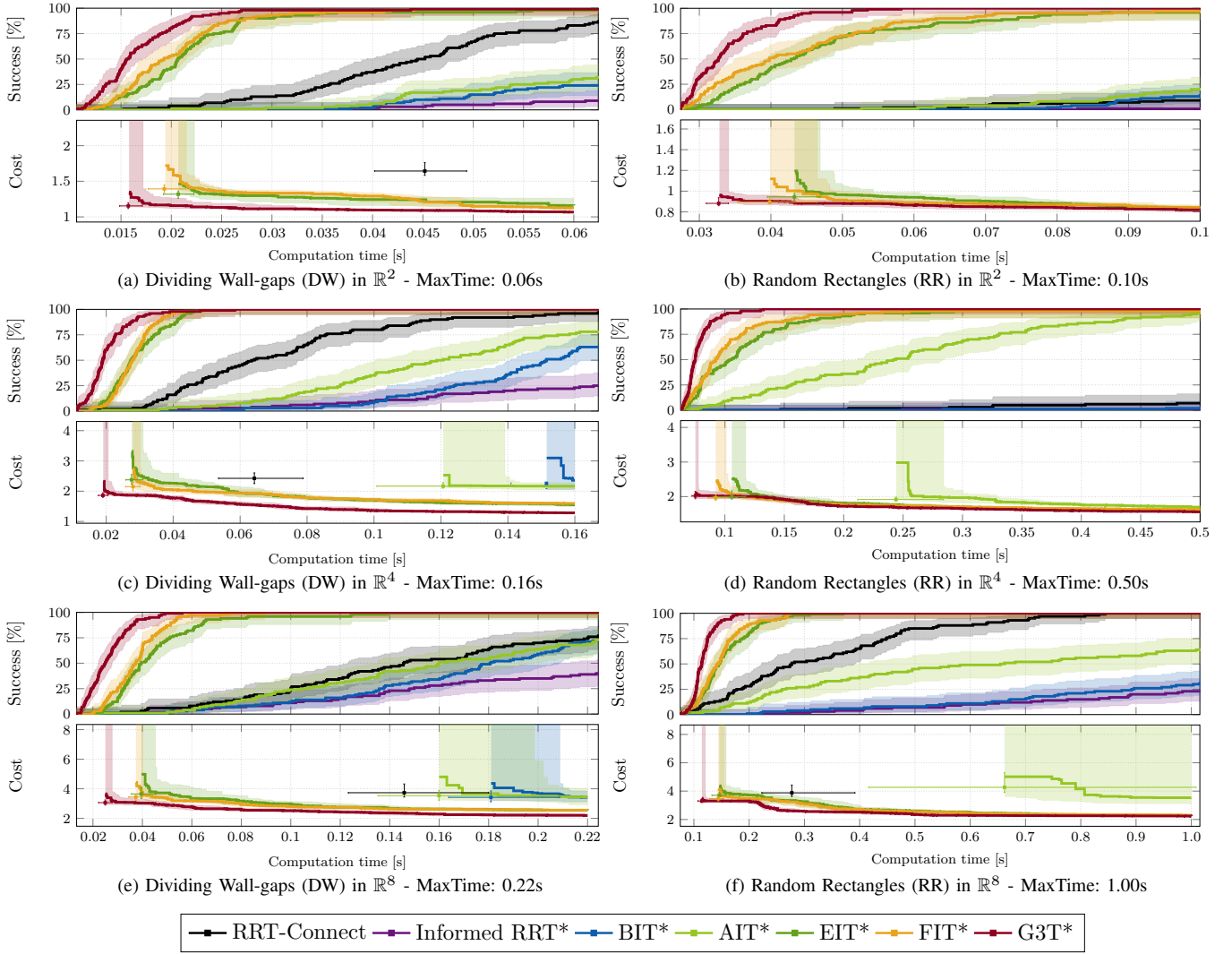
To benchmark the performance of motion planning algorithms, this study utilized the Planner-Arena benchmark database~\citeblue{moll2015benchmarking}, along with Planner Developer Tools (PDT)~\citeblue{gammell2022planner} and MoveIt~\citeblue{gorner2019moveit}. The proposed G3T* algorithm was compared against SOTA methods in both simulated random scenarios (Fig.~\ref{fig: testEnv}) and real-world manipulation problems (Fig.~\ref{fig: Realresult}). The benchmark included multiple algorithms sourced from the Open Motion Planning Library (OMPL)~\citeblue{sucan2012open}, such as RRT-Connect, Informed RRT*, BIT*, AIT*, EIT*, and FIT*. 
The tests covered simulated environments ranging in dimensionality from $\mathbb{R}^2$ to $\mathbb{R}^8$, focusing primarily on path length (cost) minimization. Key parameters were standardized across planners: the RGG constant $\eta$ was set to 1.001, and the rewire factor to 1.2. For RRT-based planners, a goal bias of 5\% was implemented, with maximum edge lengths set at 0.5, 1.25, and 3.0 for $\mathbb{R}^2$, $\mathbb{R}^4$, and $\mathbb{R}^8$, respectively. The level-up threshold $\noImpThreshold$ for historical sampling is set at three. All planners employed Euclidean distance as the primary metric and effort as heuristics. G3T* leverages tree-based grafting to reconnect invalid edges during the search. It then uses a greedy GuILD subset approach to accelerate path optimization in critical target areas, followed by a historical distribution sampling strategy to ensure asymptotic optimality. \textcolor{black}{The implementation of G3T* planner into OMPL framework is available at: }\href{https://github.com/Liding-Zhang/ompl_g3t.git}{\textcolor{gray}{https://github.com/Liding-Zhang/ompl\_g3t.git}}.
\begin{table}[t]
\caption{Benchmarks evaluation comparison (100 runs) (Fig.~\ref{fig: result})}
\centering
\resizebox{0.485\textwidth}{!}{
\begin{tabular}{lccccccc}
 \toprule

 & \multicolumn{3}{c}{${\text{FIT* (SOTA)}}$} & \multicolumn{3}{c}{$\textcolor{purple}{\text{G3T* (ours)}}$} &\multirow{2}*{\large$t^\textit{med}_\textit{init}\color{purple}\Uparrow$ (\%)}\\
\cmidrule(lr){2-4} \cmidrule(lr){5-7}
    &$t^\textit{med}_\textit{init}$ &$c^\textit{med}_\textit{init}$ &$c^\textit{med}_\textit{final}$ &$t^\textit{med}_\textit{init}$ &$c^\textit{med}_\textit{init}$ &$c^\textit{med}_\textit{final}$ \\
    
 \midrule

    \textcolor{purple}{$\text{DW}-\mathbb{R}^2$}     &0.0193 &1.3932 &1.1244 &0.0157 &\textcolor{purple}{1.1541} &1.0670  &18.65\\
    \textcolor{purple}{$\text{DW}-\mathbb{R}^4$}&0.0280  &2.1506  &1.5960 &\textcolor{purple}{0.0190} &{1.8604} &\textcolor{purple}{1.2844} &\textcolor{purple}{32.14}  \\
    \textcolor{purple}{$\text{DW}-\mathbb{R}^8$}   &0.0374   &3.4391   &2.5419 &\textcolor{purple}{0.0250} &{3.0613} &{2.1880}  &\textcolor{purple}{33.15}  \\
\midrule
    $\text{RR}-\mathbb{R}^2$   &0.0398   &0.8950  &0.8401 &0.0319 &0.8822 &0.8191 &19.85  \\
    $\text{RR}-\mathbb{R}^4$   &0.0923   &1.9647   &1.6257 &0.0721 &{2.0197} &\textcolor{purple}{1.5579}  &{21.88}  \\
    $\text{RR}-\mathbb{R}^{8}$   &{0.1471} &{3.4760} &{2.3011} &\textcolor{purple}{0.1104} &{3.3030} &{2.2335}  &\textcolor{purple}{24.95}\\

\bottomrule
\end{tabular}} \label{tab:benchmark}
\vspace{-1.3em} 
\end{table}
\vspace{-1.0em}
\subsection{Simulation Experimental Tasks}\label{subsec:experi}
The planners were evaluated across two distinct benchmarks in three domains: $\mathbb{R}^2$, $\mathbb{R}^4$, and $\mathbb{R}^{8}$. \textcolor{black}{In the first scenario, a constrained environment resembling dividing wall gaps (DW) includes multiple  corridors with spacious regions, testing the ability of motion planners to navigate multi-narrow spaces while identifying an optimal path to the goal (Fig.~\ref{fig: testEnv}a). Each planner was tested over 100 runs with computation times for each optimal planner detailed in the labels, using random seeds. The overall success rates and median path lengths for all planners are presented in Fig.~\ref{fig: result}a, Fig.~\ref{fig: result}c, and Fig.~\ref{fig: result}e. Results indicate that G3T* quickly finds an initial solution across dimensions with faster path convergence.}

\vspace{-0.13em}
In the second test scenario, random widths were assigned to \textit{axis-aligned hyperrectangles}, which were generated arbitrarily within the \textit{$\mathcal{C}$-space} (Fig.~\ref{fig: testEnv}b). For each dimension of the \textit{$\mathcal{C}$-space}, random rectangle problems were created, and each planner underwent 100 runs for every instance. Figures~\ref{fig: result}b, \ref{fig: result}d, and \ref{fig: result}f demonstrate that the proposed method achieved the highest success rates and the lowest median path costs within the given computation time, compared to the other planners. This suggests that G3T* can optimize paths more quickly and converge faster through grafting. G3T* outperformed the other planners and was able to quickly find an initial solution.

As observed in benchmark Table~\ref{tab:benchmark}, the notable improvement (marked in red) in the median initial times  over 100 runs is across different benchmark scenarios, with performance trends correlating with dimensionality. \textcolor{black}{In the $\text{DW}-\mathbb{R}^4$ scenario, G3T* achieves a lower initial median time (0.019s, calculated over 100 trials) compared to FIT* (0.028s). Furthermore, in higher-dimensional scenarios such as $\text{RR}-\mathbb{R}^8$ and $\text{DW}-\mathbb{R}^8$, G3T* consistently outperforms FIT* by achieving lower initial median times with higher success rate. Specifically, in the $\text{DW}-\mathbb{R}^8$ scenario, G3T* demonstrates a notable initial median time improvement of approximately 33.15\% over FIT*.}

Overall, G3T* shows faster initial path convergence compared to the SOTA planner, thereby enhancing the search efficiency of sampling-based path planning algorithms.
\vspace{-1.0em}
\subsection{Real-world Path Planning Tasks}\label{subsec:realExpri}
To evaluate the algorithm's performance in real-world scenarios, kitchen model pepper problem experiments (Fig.~\ref{fig: Realresult}) are conducted on an 8-DoF base manipulator (DARKO) to demonstrate the efficiency and scalability of G3T*, compared to three SOTA path planning algorithms: Batch Informed Trees (BIT*)~\citeblue{gammell2015batch, gammell2020batch}, Effort Informed Trees (EIT*)~\citeblue{strub2022adaptively}, and Flexible Informed Trees (FIT*)~\citeblue{Zhang2024adaptive}.
We compare G3T* with EIT* and FIT* in robot manipulation tasks to evaluate performance in terms of convergence to optimal solution cost and success rate over 30 runs. The \textit{Kitchen-pepper} experiment is conducted on the DARKO robot in a cluttered environment with narrow spaces. A collision-free path is required from the lower-shelf section of the kitchen model (start config.) to the upper-left cooking pan area (goal region). G3T* demonstrates effective grafting for path convergence with greedy GuILD sets for path optimization throughout the experimental tasks. 
%

Each planner was allocated 1.0 seconds to solve this confined, limited space pull-out and manipulate problem. Across 30 trials,  EIT* managed a 33.33\% success rate with a median solution cost of 20.7864. FIT* had a 70\% success rate with a median solution cost of 22.6911. G3T* achieved an 80\% success rate with a median solution cost of 17.9354.

In summary, compared to EIT* and FIT*, G3T* achieves the best performance in both finding the initial solution and converging to the optimal solution.

%% file: sections/sec7_conclusion.tex
\section{Conclusion and Future Work}
In this paper, we introduce the Greedy GuILD Grafting Trees (G3T*), a novel path-planning algorithm that employs a tree-based grafting method for bidirectional search. G3T* defines a common neighbor set for the invalid edges' endpoints and forms new optimal edge-pairs to continue the search. For fast path optimization, G3T* defines a greedy GuILD approach based on the maximal admissible cost of the front-back path as a two-subdomain problem. G3T* utilizes historical distribution sampling to prevent premature convergence to local minima and adjusts the sampling level based on both current and historical cost improvements.
By historical sampling, the probabilistic completeness and asymptotic optimality were guaranteed. Simulations across varying dimensions and real-world experiments validate that our algorithm achieves faster initial convergence and shorter path lengths than the benchmarks.

\textcolor{black}{Future work could integrate safety constraints for human-interactive scenarios, and utilize single instruction/multiple data parallelism methods~\citeblue{Wilson2024fcit} to improve motion planning.}